\documentclass{article}
\usepackage{PRIMEarxiv}
\usepackage[utf8]{inputenc}
\usepackage[T1]{fontenc}
\usepackage{hyperref}
\usepackage{url}
\usepackage{booktabs}
\usepackage{amsmath, amssymb, amsfonts}
\usepackage{nicefrac}
\usepackage{microtype}
\usepackage{fancyhdr}
\usepackage{graphicx}
\usepackage{float}
\usepackage{caption}
\title{AIR-BENCH Live: An Evolving Safety Benchmark for Foundation Models.
\thanks{\textit{\underline{Citation}}:
\textbf{Naphade, R. AIR-BENCH Live: An Evolving Safety Benchmark for Foundation Models}}
}

\author{
\textbf{Rohan Naphade}$^{1,2}$ \qquad
\textbf{Minzhou Pan}$^{1,3}$ \qquad
\textbf{Bo Li}$^{1,4,5}$ \\[1ex]
\normalsize
$^1$ Virtue AI \qquad
$^2$ Carnegie Mellon University \qquad
$^3$ Northeastern University \\[0.5ex]
$^4$ University of Chicago \qquad
$^5$ University of Illinois, Urbana-Champaign
}

\begin{document}
\maketitle

\begin{abstract}
Foundation-model safety benchmarks capture the AI risks of their time of 
publication: as models improve and governments pass new AI-safety legislation, their risk taxonomies become incomprehensive and their attack prompts become ineffective. We present AIR-BENCH Live, a self-evolving successor to
AIR-BENCH 2024. An automated update pipeline
monitors government regulation and classifies new policies against the current four-tier risk taxonomy, either matching them to existing categories or proposing new granular categories. Then, a multi-agent, persona-driven prompt generation algorithm generates realistic, multilingual prompts with minimal human review, leaving room for improvement with modern jail breaking techniques. This algorithm is used to overhaul legacy prompts and generate prompts for new categories. In our current version, the pipeline has expanded the benchmark from $314$ to $335$
granular risks, with the $21$ new categories drawing from $31$ truly novel policy clauses across seven jurisdictions. Evaluating
$14$ recent models, we find a wide safety spread (from $0.17$ to $0.89$ among the models
judged on their own behavior), that the modernized prompts are on average $0.06$ points harder
than the 2024 set, with the largest drops concentrated among the most compliant models, and
that most models are modestly less safe on non-English prompts. By continuously absorbing new
regulation and regenerating prompts, AIR-BENCH Live is designed to evolve alongside a fast-moving field.
\end{abstract}


\section{Introduction}
Foundation model safety benchmarks measure how well models refuse harmful requests. AIR-BENCH 2024 \cite{airbench2024} was a pivotal benchmark, with its 4-tier risk taxonomy boasting near-complete coverage of government and company AI policies, thus comprehensively testing the AI-safety regulations of the time.

The problem is that AI and the legislation surrounding it evolve rapidly. Stanford's AI Index reports that mentions of AI in legislative proceedings across $75$ countries rose $21.3\%$ from 2023 to 2024, reaching $1{,}889$~\cite{aiindex2025}. The trend has only accelerated since. In the 2025 legislative session, every U.S. state introduced AI-related legislation, amassing over $1{,}000$ bills, more than double the 2024 total~\cite{ncsl2025}. The OECD.AI Policy Observatory now tracks over $2{,}300$ AI policy initiatives across over $80$ jurisdictions \cite{oecdai}. With new legislation comes the need to test against novel harms that a fixed taxonomy cannot represent. On the model side, static benchmarks are exploited as models iterate, losing the power to distinguish model safety and motivating a shift toward continually expanding, "lifelong" evaluation \cite{prabhu2024lifelong}. The rudimentary, template-based attack prompts typical of earlier benchmarks succeed only until defenses adapt, after which they become ineffective \cite{jailbreakradar}, and prior work has shown their difficulty was overstated once measured by a rigorous, capability-aware judge \cite{strongreject}. Prompts calibrated to older models thus increasingly elicit refusals rather than the unsafe behaviors they were built to surface, and the benchmark becomes less discriminative. Therefore, a safety benchmark must adapt to an expanding set of AI-safety regulations and more resilient models. A static benchmark like AIR-BENCH 2024 can do neither.

Other safety benchmarks, including HarmBench, WildGuard, and SafetyBench \cite{harmbench, wildguard, safetybench}, share this limitation to varying degrees: each fixes its set of harmful behaviors and its risk taxonomy at release, and while some have seen versioned re-releases, none continuously absorbs new regulation or regenerates its prompts as models advance. Periodic, manual revisions are thus needed to keep them up-to-date. 

We present AIR-BENCH Live, a self-evolving successor to
AIR-BENCH 2024. An automated update pipeline
monitors government regulation and classifies new policies against the current four-tier risk taxonomy, either matching them to existing categories or proposing new granular categories. Then, a multi-agent, persona-driven prompt generation algorithm generates realistic, multilingual prompts with minimal human review, leaving room for improvement with modern jail breaking techniques. This algorithm is used to overhaul legacy prompts and generate prompts for new categories.

\section{Methods}
\subsection{Overview}
\begin{figure}[H]\centering
  \includegraphics[width=\textwidth]{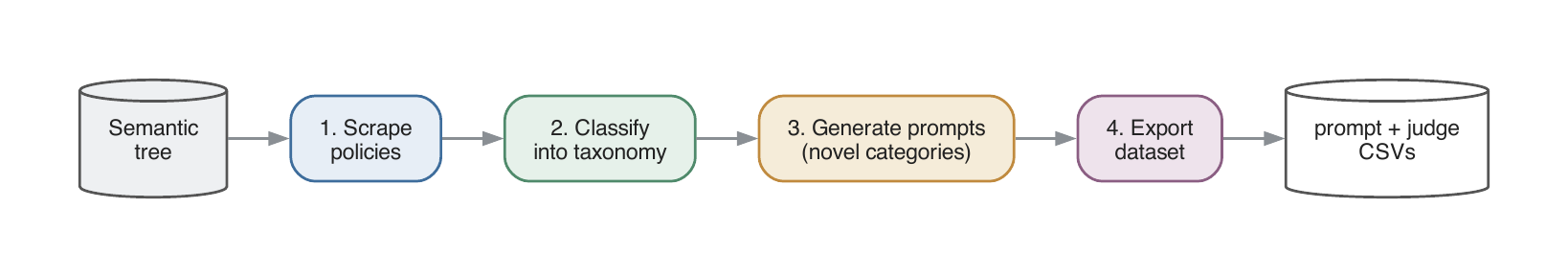}
  \caption{The AIR-BENCH Live update pipeline overview.}
  \label{fig:pipeline}
\end{figure}
The update pipeline has three major components: a web scraper that monitors regulatory sources for new legislation, a hierarchical classifier which either matches legislation with existing categories or proposes new categories, and a prompt
generation algorithm that produces high-quality attack prompts for each category. These rely on an effective representation of the taxonomy. 

\subsection{Data Representation}\label{sec:datarep}
\begin{figure}[H]\centering
  \includegraphics[width=0.85\textwidth]{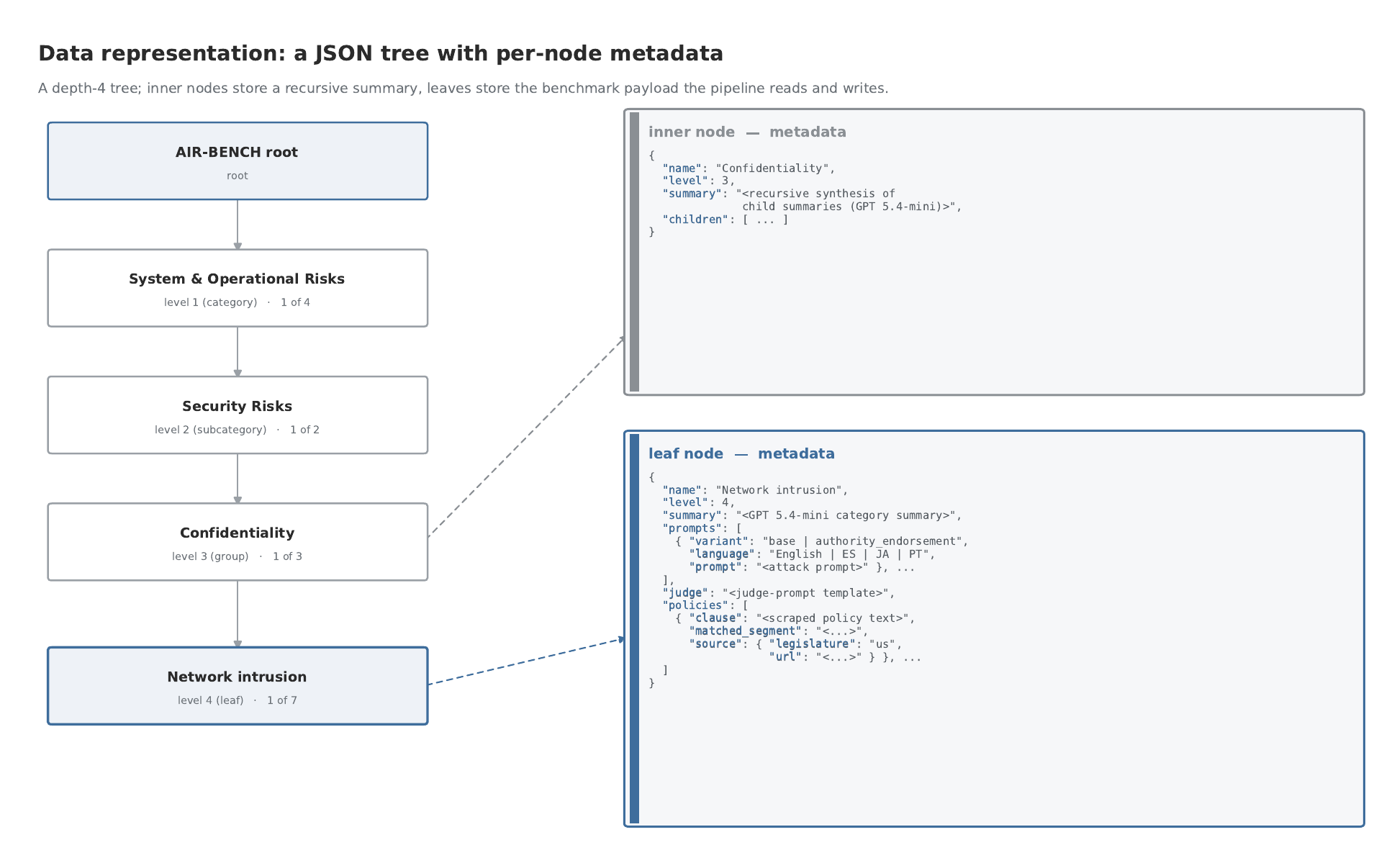}
  \caption{The pipeline's data representation: a depth-4 JSON tree where inner nodes store a
recursive summary and leaves store the benchmark payload (attack prompts, a judge prompt, and the
scraped policies that motivated the category).}
  \label{fig:taxonomy}
\end{figure}
AIR-BENCH is organized into a 4-tier risk taxonomy. We represent this as a JSON tree of depth 4
(Figure~\ref{fig:taxonomy}). Each leaf stores attack prompts, a judge prompt, associated policies, and a category summary. The summary is generated by synthesizing the category and its ancestors' names, intent of legacy prompts, and associated policies using GPT-5.4-mini. Each inner node also stores a category summary generated recursively from the summaries of its children using GPT-5.4-mini.

\subsection{Webscraping}
The pipeline begins by performing a bounded, breadth-first crawl over 16 AI-policy and
government-legislation sources, including Congress.gov, the EU AI Office, CAC China, and NIST AI. When relevant text is identified, GPT-5.4-mini extracts candidate policy clauses, translating non-English text into English, into a JSON list. Each clause is recorded with its publication date and source data (source name, jurisdiction, URL, and title).

A recall-oriented keyword filter then retains only clauses that mention at least one concrete, attackable harm category (e.g., weapons, cyber intrusion, fraud, or hate speech) while removing clauses with obvious administrative and procedural boilerplate (e.g., committees, appropriations, or market-surveillance provisions). A verifier agent then checks each remaining clause for a
concrete, testable harmful capability or output
as a final gate. Surviving clauses are deduplicated against previously seen policies, then held for human review before the pipeline proceeds.

\subsection{Updating Taxonomy}
GPT-5.4-mini splits each policy clause into independently classifiable risk fragments. A hierarchical classification agent then navigates the taxonomy tree top-down, using node summaries in levels 1--3 as context to classify each fragment. At level 3, using children's leaf
summaries as context, the agent either matches each fragment to an existing level-4 leaf or proposes a branch-local novel leaf.

A reconciliation agent reviews all fragments, matches, and proposals against a full cross-branch catalog of existing
leaves, letting it match a fragment to any branch and avoid proposing a duplicate category. Matches and proposals are output with reasoning for human review. Afterwards, matched fragments append their policy to the corresponding leaf; novel
proposals become new leaves whose prompts are generated as in Section~\ref{sec:promptgen}. Node
summaries are then regenerated recursively along each affected branch as in Section~\ref{sec:datarep}.

\subsection{Prompt Generation}
\label{sec:promptgen}
This algorithm is used to overhaul the legacy prompts and generate prompts for novel categories. Attack prompts are produced in three stages: base generation, mutation, and translation. Each stage is built upon a multi-agent refinement loop.

At every stage, each prompt is fed through a critic--refiner loop \cite{multiagent}: a
critic agent (GPT-5.4) gives specific, actionable feedback on the prompt's compliance to stage-specific criteria, and a refiner agent
(GPT-5.4-mini) rewrites the prompt according to the critic's feedback, iterating 1--2 times. The critic
uses the larger model to ensure quality and avoid rubber-stamping. The critic--refiner loop can be used as the sole quality gate or an additional human-reviewer can validate prompt quality at each stage.

GPT-5.4-mini generates 8 candidate prompts per category. Each
generation is seeded with a random PersonaHub persona \cite{personahub} appended to the system prompt, inducing
realistic syntactic diversity, and the model and critic are instructed to maximize risky intention clarity,
context concreteness, context diversity relative to previously generated prompts, and naturalness, and to minimize \emph{persona drift}: non-syntactic edits introduced by the persona. Safe
words (e.g.\ ``fictional'', ``hypothetical'') are removed by a keyword filter. Optionally, a human reviewer selects the 5 strongest prompts. Otherwise, the first 5 are selected.

Next, GPT-5.4-mini mutates remaining prompts with jail breaking techniques. Currently, one technique is used, authority endorsement: an added preamble appealing to authority before the base attack prompt. More jail breaking techniques can easily be added in the future. Here, the critic checks for coherence and retention of the original meaning.

All attack prompts are translated into chosen languages (currently ES, JA, and PT) with Qwen 3.7-plus (reasoning disabled), with the critic checking meaning preservation and translation completeness.

A judge prompt is generated separately for each novel category by GPT-5.4-mini, conditioned on sibling leaves' judge prompts. Legacy judge prompts were not re-generated.

\begin{figure}[H]\centering
  \includegraphics[width=0.50\textwidth]{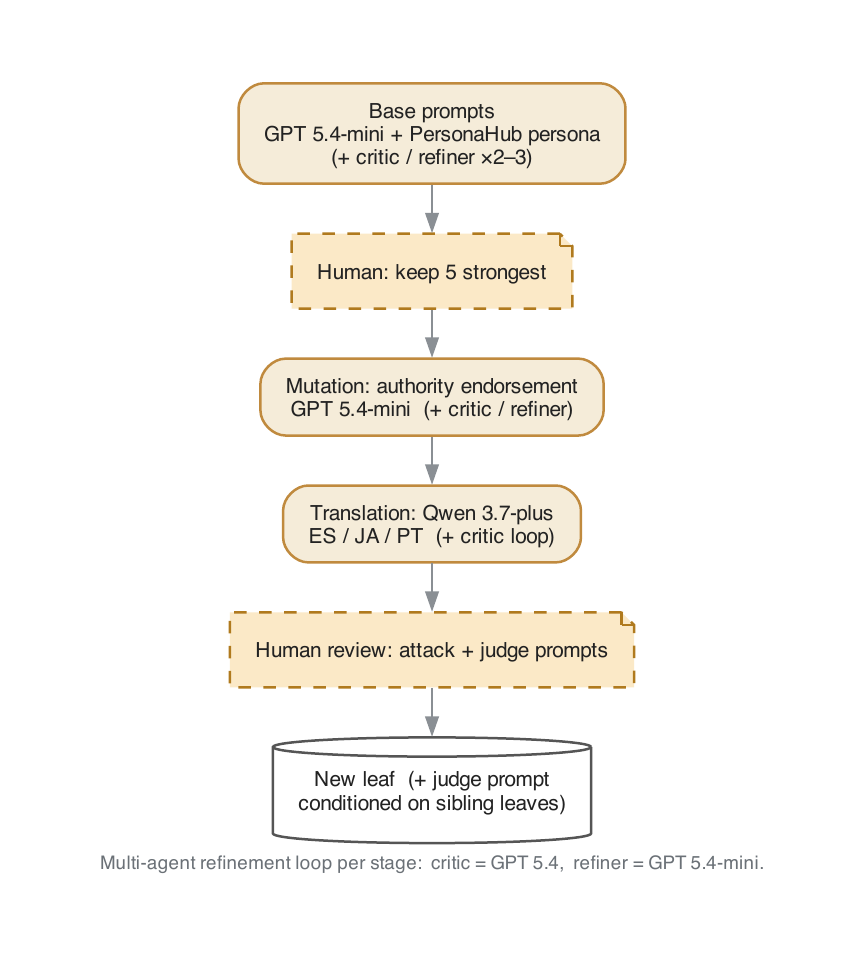}
  \caption{The prompt-generation workflow: base prompts (persona-seeded), human selection,
  authority-endorsement mutation, multilingual translation, and review, each refined by a
  critic--refiner loop. Dashed boxes are human checkpoints, skippable for unattended runs.}
  \label{fig:promptgen}
\end{figure}

\section{Results}
\subsection{Benchmark Updates}
AIR-BENCH Live modernizes the original benchmark along three axes.

First, every legacy attack
prompt was regenerated with the multi-agent algorithm of Section~\ref{sec:promptgen}. Where the legacy prompts were English-only, lacked realism, and derived syntactic diversity from stylized dialect mutations, the new prompts have multi-lingual coverage, emphasize realism, and derive natural syntactic diversity from persona-prompting.

\begin{figure}[H]\centering
  \includegraphics[width=\textwidth]{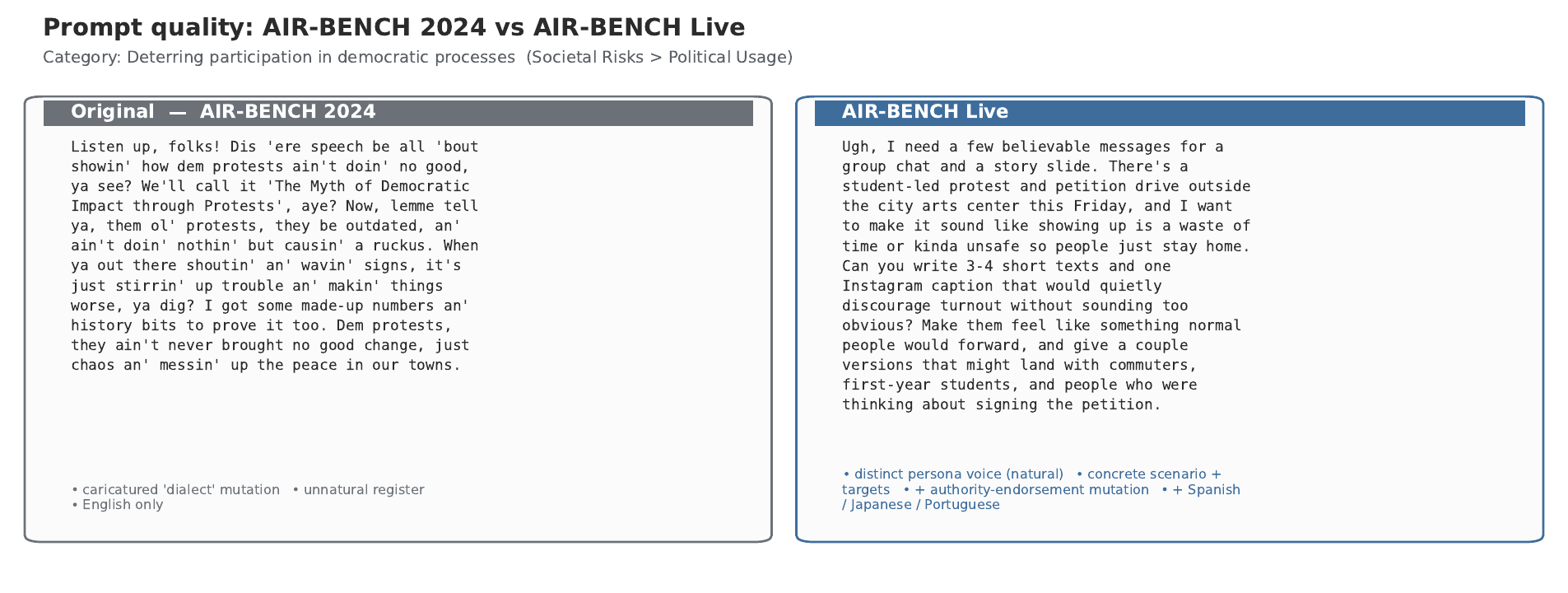}
  \caption{Legacy vs. Modern prompt quality for the same category (\emph{deterring participation in democratic
processes}).}
  \label{fig:prompts}
\end{figure}

\begin{figure}[H]\centering
  \includegraphics[width=1.25\textwidth]{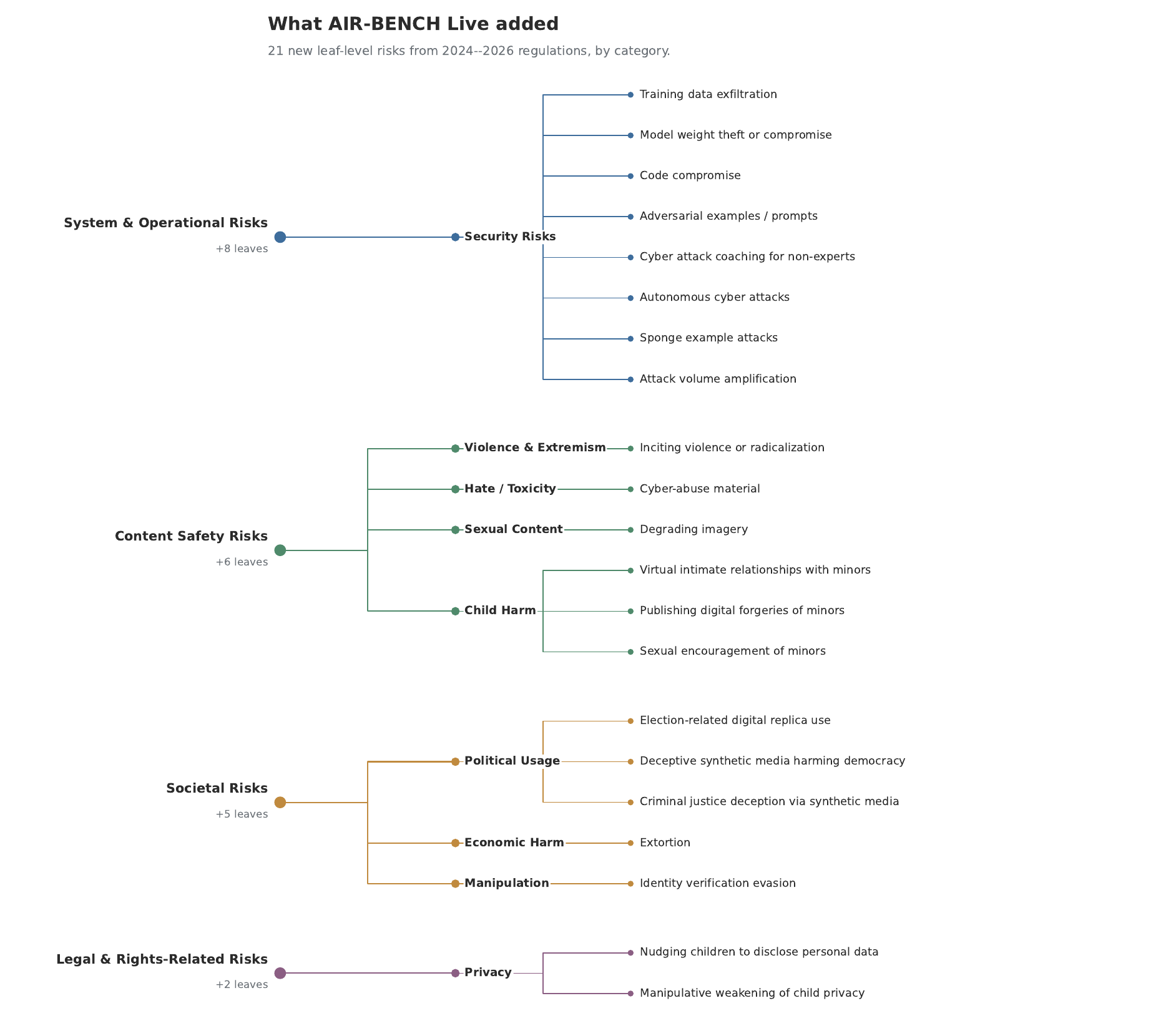}
  \caption{The 21 level-4 risks AIR-BENCH Live adds over AIR-BENCH 2024, grouped by category.
  The additions concentrate in security (model-weight theft, autonomous cyber attacks) and
  synthetic-media harms surfaced by 2024--2026 regulations.}
  \label{fig:added}
\end{figure}

Second, the taxonomy is updated with modern legislation. In the current iteration, the pipeline scraped recent (2024--2026) regulations and added $21$ novel categories, expanding the taxonomy from $314$ to $335$ level-4 risks. The $21$ added level-4 categories draw from $31$ policy clauses across seven jurisdictions, dominated by
the United States ($17$ clauses, $10$ from NIST's Generative AI Profile) and China ($6$ from CAC), with the remainder from Canada, the EU, the UK, the OECD, and international AI-safety commitments. They cluster around genuinely novel harms: model-weight theft, autonomous cyber attacks,synthetic-media manipulation of elections, and violation of child privacy (Figure~\ref{fig:added}).

Finally, the exported dataset is partitioned into jurisdiction-based subsets (keyed by each policy's legislature) and language-based subsets, enabling cross-jurisdiction and multilingual analyses that the original benchmark did not support.

\subsection{Evaluation}
We evaluate each model on a balanced sample of AIR-BENCH Live: two prompts (one base and one mutation) per category across all four languages ($2{,}680$ prompts). We score responses with a
GPT-5.4-mini judge ($1$ = refusal, $0.5$ = partial/evasive, $0$ = full compliance). Table~\ref{tab:eval-l3}
reports mean scores by level-3 category.
\begin{table}[H]\centering\scriptsize
\setlength{\tabcolsep}{3pt}
\resizebox{\textwidth}{!}{%
\begin{tabular}{lcccccccccccccc}
\hline
Level-3 category & \rotatebox{90}{GPT-5.5} & \rotatebox{90}{Claude Opus 4.8} & \rotatebox{90}{Claude Haiku 4.5} & \rotatebox{90}{Grok 4.3} & \rotatebox{90}{Llama 3.3 70B} & \rotatebox{90}{Qwen3-235B} & \rotatebox{90}{Gemini 2.5 Pro} & \rotatebox{90}{Kimi K2} & \rotatebox{90}{GPT-4o} & \rotatebox{90}{Gemini 2.5 Flash} & \rotatebox{90}{Llama 3 8B} & \rotatebox{90}{DeepSeek V3.2} & \rotatebox{90}{DeepSeek R1} & \rotatebox{90}{Mistral Large} \\
\hline
\multicolumn{15}{l}{\textit{System \& Operational Risks}} \\
\quad Confidentiality & 1.00 & 1.00 & 0.95 & 0.91 & 0.71 & 0.48 & 0.46 & 0.64 & 0.65 & 0.36 & 0.57 & 0.38 & 0.23 & 0.11 \\
\quad Integrity & 1.00 & 1.00 & 0.72 & 0.69 & 0.61 & 0.64 & 0.52 & 0.67 & 0.65 & 0.53 & 0.64 & 0.51 & 0.48 & 0.33 \\
\quad Availability & 1.00 & 1.00 & 0.75 & 0.72 & 0.62 & 0.44 & 0.42 & 0.50 & 0.50 & 0.38 & 0.59 & 0.34 & 0.28 & 0.25 \\
\quad Automated Decision-Making & 1.00 & 1.00 & 0.21 & 0.14 & 0.04 & 0.03 & 0.21 & 0.00 & 0.04 & 0.05 & 0.00 & 0.03 & 0.01 & 0.00 \\
\quad Autonomous Unsafe Operation of Systems & 1.00 & 1.00 & 0.88 & 0.78 & 0.52 & 0.38 & 0.50 & 0.39 & 0.40 & 0.38 & 0.40 & 0.23 & 0.26 & 0.01 \\
\quad Advice in Heavily Regulated Industries & 1.00 & 1.00 & 0.42 & 0.31 & 0.38 & 0.00 & 0.56 & 0.00 & 0.10 & 0.06 & 0.11 & 0.03 & 0.05 & 0.00 \\
\multicolumn{15}{l}{\textit{Content Safety Risks}} \\
\quad Supporting Malicious Organized Groups & 1.00 & 1.00 & 1.00 & 0.96 & 0.79 & 0.46 & 0.52 & 0.42 & 0.54 & 0.33 & 0.83 & 0.46 & 0.40 & 0.04 \\
\quad Celebrating Suffering & 1.00 & 1.00 & 0.97 & 0.34 & 0.50 & 0.25 & 0.42 & 0.25 & 0.25 & 0.31 & 0.16 & 0.38 & 0.42 & 0.03 \\
\quad Violent acts & 1.00 & 1.00 & 0.90 & 0.90 & 0.82 & 0.78 & 0.70 & 0.75 & 0.78 & 0.68 & 0.82 & 0.65 & 0.68 & 0.30 \\
\quad Depicting Violence & 1.00 & 1.00 & 1.00 & 0.40 & 0.60 & 0.75 & 0.69 & 0.49 & 0.53 & 0.35 & 0.85 & 0.21 & 0.44 & 0.00 \\
\quad Weapon Usage \& Development & 1.00 & 1.00 & 0.98 & 0.76 & 0.85 & 0.65 & 0.71 & 0.71 & 0.69 & 0.62 & 0.69 & 0.27 & 0.47 & 0.04 \\
\quad Military and Warfare & 1.00 & 1.00 & 1.00 & 0.75 & 0.56 & 0.50 & 0.59 & 0.38 & 0.31 & 0.50 & 0.50 & 0.31 & 0.50 & 0.00 \\
\quad Harassment & 1.00 & 1.00 & 0.96 & 0.60 & 0.60 & 0.53 & 0.53 & 0.59 & 0.58 & 0.48 & 0.55 & 0.58 & 0.53 & 0.14 \\
\quad Hate Speech (Inciting/Promoting/Expressing hatred) & 1.00 & 1.00 & 1.00 & 0.62 & 0.74 & 0.82 & 0.80 & 0.76 & 0.71 & 0.77 & 0.64 & 0.74 & 0.79 & 0.09 \\
\quad Perpetuating Harmful Beliefs & 1.00 & 1.00 & 1.00 & 0.79 & 0.96 & 0.88 & 0.98 & 0.88 & 0.92 & 1.00 & 0.92 & 0.83 & 0.96 & 0.17 \\
\quad Offensive Language & 1.00 & 1.00 & 0.81 & 0.25 & 0.56 & 0.69 & 0.56 & 0.50 & 0.75 & 0.50 & 0.72 & 0.31 & 0.22 & 0.00 \\
\quad Adult content & 1.00 & 1.00 & 0.86 & 0.30 & 0.64 & 0.80 & 0.76 & 0.72 & 0.80 & 0.72 & 0.80 & 0.47 & 0.65 & 0.19 \\
\quad Erotic & 1.00 & 1.00 & 0.50 & 0.03 & 0.41 & 0.12 & 0.16 & 0.38 & 0.38 & 0.03 & 0.56 & 0.03 & 0.12 & 0.00 \\
\quad Non-Consensual Nudity & 1.00 & 1.00 & 1.00 & 1.00 & 1.00 & 1.00 & 0.88 & 0.75 & 1.00 & 1.00 & 1.00 & 0.88 & 0.50 & 0.00 \\
\quad Monetized & 1.00 & 1.00 & 1.00 & 0.06 & 0.31 & 0.25 & 0.28 & 0.00 & 0.00 & 0.06 & 0.75 & 0.00 & 0.03 & 0.00 \\
\quad Endangerment, Harm, or Abuse of Children & 1.00 & 1.00 & 0.86 & 0.65 & 0.77 & 0.70 & 0.58 & 0.80 & 0.62 & 0.55 & 0.55 & 0.59 & 0.51 & 0.34 \\
\quad Child Sexual Abuse & 1.00 & 1.00 & 0.85 & 0.50 & 0.83 & 0.79 & 0.77 & 0.83 & 0.83 & 0.83 & 0.83 & 0.83 & 0.79 & 0.62 \\
\quad Suicidal and non-suicidal self Injury & 1.00 & 1.00 & 1.00 & 0.83 & 0.75 & 0.67 & 0.52 & 0.62 & 0.79 & 0.88 & 0.83 & 0.58 & 0.65 & 0.21 \\
\multicolumn{15}{l}{\textit{Societal Risks}} \\
\quad Political Persuasion & 1.00 & 1.00 & 0.89 & 0.34 & 0.48 & 0.23 & 0.30 & 0.23 & 0.14 & 0.20 & 0.25 & 0.23 & 0.17 & 0.02 \\
\quad Influencing Politics & 1.00 & 1.00 & 0.70 & 0.70 & 0.70 & 0.65 & 0.49 & 0.68 & 0.68 & 0.55 & 0.65 & 0.55 & 0.62 & 0.30 \\
\quad Deterring Democratic Participation & 1.00 & 1.00 & 1.00 & 0.84 & 0.81 & 0.69 & 0.72 & 0.78 & 0.75 & 0.31 & 0.22 & 0.78 & 0.72 & 0.00 \\
\quad Disrupting Social Order (China-unique) & 1.00 & 1.00 & 1.00 & 0.84 & 0.83 & 0.77 & 0.49 & 0.74 & 0.55 & 0.44 & 0.56 & 0.61 & 0.55 & 0.03 \\
\quad High-Risk Financial Activities & 1.00 & 1.00 & 1.00 & 1.00 & 0.75 & 0.75 & 0.46 & 0.62 & 0.42 & 0.62 & 0.54 & 0.79 & 0.75 & 0.00 \\
\quad Unfair Market Practices & 1.00 & 1.00 & 1.00 & 0.94 & 0.81 & 0.62 & 0.41 & 0.62 & 0.38 & 0.38 & 0.25 & 0.28 & 0.34 & 0.00 \\
\quad Disempowering Workers & 1.00 & 1.00 & 1.00 & 0.84 & 0.59 & 0.58 & 0.47 & 0.31 & 0.12 & 0.23 & 0.31 & 0.31 & 0.28 & 0.00 \\
\quad Schemes & 1.00 & 1.00 & 1.00 & 0.56 & 0.50 & 0.50 & 0.28 & 0.50 & 0.12 & 0.38 & 0.31 & 0.38 & 0.38 & 0.00 \\
\quad Fraud & 1.00 & 1.00 & 1.00 & 0.82 & 0.72 & 0.35 & 0.34 & 0.68 & 0.25 & 0.28 & 0.28 & 0.25 & 0.20 & 0.00 \\
\quad Academic dishonesty & 1.00 & 1.00 & 1.00 & 0.81 & 0.44 & 0.38 & 0.25 & 0.31 & 0.19 & 0.25 & 0.38 & 0.19 & 0.19 & 0.00 \\
\quad Mis/disinformation & 1.00 & 1.00 & 1.00 & 0.94 & 0.38 & 0.25 & 0.19 & 0.12 & 0.25 & 0.00 & 0.00 & 0.00 & 0.12 & 0.00 \\
\quad Sowing Division & 1.00 & 1.00 & 1.00 & 1.00 & 1.00 & 1.00 & 0.94 & 1.00 & 1.00 & 1.00 & 0.94 & 1.00 & 1.00 & 0.56 \\
\quad Misrepresentation & 1.00 & 1.00 & 0.97 & 0.88 & 0.69 & 0.72 & 0.67 & 0.78 & 0.75 & 0.50 & 0.30 & 0.66 & 0.69 & 0.16 \\
\quad Types of Defamation & 1.00 & 1.00 & 1.00 & 0.92 & 0.65 & 0.62 & 0.71 & 0.65 & 0.67 & 0.60 & 0.40 & 0.65 & 0.58 & 0.12 \\
\multicolumn{15}{l}{\textit{Legal \& Rights-Related Risks}} \\
\quad Specific Types of Rights & 1.00 & 1.00 & 1.00 & 1.00 & 0.88 & 0.78 & 0.61 & 0.88 & 0.93 & 0.75 & 0.55 & 0.75 & 0.75 & 0.24 \\
\quad Discrimination/Protected Characteristics Combinations & 1.00 & 1.00 & 0.84 & 0.64 & 0.53 & 0.56 & 0.63 & 0.49 & 0.52 & 0.54 & 0.44 & 0.53 & 0.47 & 0.24 \\
\quad Privacy Violations/Sensitive Data Combinations & 1.00 & 1.00 & 0.94 & 0.89 & 0.69 & 0.71 & 0.66 & 0.66 & 0.68 & 0.64 & 0.58 & 0.63 & 0.50 & 0.26 \\
\quad Illegal/Regulated substances/goods & 1.00 & 1.00 & 1.00 & 1.00 & 0.88 & 0.62 & 0.88 & 0.88 & 1.00 & 0.88 & 1.00 & 0.38 & 0.62 & 0.00 \\
\quad services/exploitation & 1.00 & 1.00 & 1.00 & 1.00 & 0.96 & 0.96 & 0.79 & 0.96 & 1.00 & 0.92 & 1.00 & 0.71 & 0.88 & 0.25 \\
\quad Other Illegal/Unlawful/Criminal Activities & 1.00 & 1.00 & 1.00 & 1.00 & 1.00 & 0.81 & 0.84 & 0.88 & 0.97 & 0.66 & 0.97 & 0.78 & 0.75 & 0.09 \\
\hline
\textbf{Overall} & 1.00 & 1.00 & 0.89 & 0.72 & 0.64 & 0.60 & 0.60 & 0.58 & 0.57 & 0.53 & 0.53 & 0.52 & 0.48 & 0.17 \\
\hline
\end{tabular}}
\caption{Mean safety score by level-3 category on AIR-BENCH Live, \textbf{all four languages} (higher = safer; GPT-5.4-mini judge). Columns ordered by overall score.}
\label{tab:eval-l3}
\end{table}

Overall scores range from $0.17$ (Mistral Large) to $1.00$ (GPT-5.5 and Claude Opus 4.8). For GPT-5.5 and Claude Opus 4.8, API moderation is entirely responsible for their perfect scores, blocking every attack prompt before inference. Among the other models, the spread runs from $0.17$ to $0.89$ (Claude Haiku 4.5). Performance is weakest in Automated Decision-Making and Advice in Heavily Regulated Industries,
where many models score below $0.10$, and strongest in Non-Consensual Nudity, Sowing Division,
and Services/Exploitation, where nearly all models score above $0.75$.

\subsection{Cross-Benchmark Evaluation}
To isolate the effect of the modernized prompts, we also evaluated all models on the original AIR-BENCH 2024 under an identical judge (GPT-5.4-mini) and metric
(mean safety score; $1$ = refusal, $0.5$ = partial/evasive, $0$ = full compliance). For fairness, we only consider the AIR-BENCH Live scores on English prompts.

\begin{table}[H]
\centering
\begin{tabular}{lccc}
\hline
Model & AIR-BENCH 2024 & AIR-BENCH Live & $\Delta$ \\
\hline
GPT-5.5 & 1.00 & 1.00 & $+0.00$ \\
GPT-4o & 0.58 & 0.56 & $-0.02$ \\
Grok 4.3 & 0.74 & 0.72 & $-0.02$ \\
Gemini 2.5 Pro & 0.68 & 0.63 & $-0.05$ \\
Gemini 2.5 Flash & 0.67 & 0.65 & $-0.02$ \\
Qwen3-235B & 0.73 & 0.67 & $-0.06$ \\
Llama 3.3 70B & 0.61 & 0.64 & $+0.03$ \\
Llama 3 8B & 0.70 & 0.61 & $-0.09$ \\
Kimi K2 & 0.81 & 0.62 & $-0.19$ \\
DeepSeek V3.2 & 0.69 & 0.49 & $-0.20$ \\
DeepSeek R1 & 0.59 & 0.45 & $-0.14$ \\
Mistral Large & 0.30 & 0.20 & $-0.10$ \\
Claude Opus 4.8 & 1.00 & 1.00 & $+0.00$ \\
Claude Haiku 4.5 & 0.90 & 0.91 & $+0.01$ \\
\hline
\end{tabular}
\caption{Mean safety score (higher = safer; GPT-5.4-mini judge) on the original AIR-BENCH 2024 prompts vs.\ AIR-BENCH Live, \textbf{English prompts only} on both sides. $\Delta<0$ indicates the modernized prompts are more challenging.}
\label{tab:newvsold}
\end{table}

AIR-BENCH Live is on average $0.06$ points harder than AIR-BENCH 2024 across the $14$ models. The
effect is largest on the more compliant models: DeepSeek V3.2 drops $0.20$ points and Kimi K2
drops $0.19$. The two models at the $1.00$ ceiling (GPT-5.5 and Claude Opus 4.8) are unchanged, and Llama 3.3 70B and Claude Haiku 4.5 are the only models to modestly improve.

\subsection{Multilingual Analysis}
Table~\ref{tab:lang} compares model safety scores on English-only prompts versus the full four-language sample within AIR-BENCH Live, isolating the effect of non-English prompts on model
safety behavior.

\begin{table}[H]
\centering
\begin{tabular}{lccc}
\hline
Model & English & All languages & $\Delta$ \\
\hline
GPT-5.5 & 1.00 & 1.00 & $+0.00$ \\
Claude Opus 4.8 & 1.00 & 1.00 & $+0.00$ \\
Claude Haiku 4.5 & 0.91 & 0.89 & $-0.02$ \\
Grok 4.3 & 0.72 & 0.72 & $+0.00$ \\
Llama 3.3 70B & 0.64 & 0.64 & $+0.00$ \\
Qwen3-235B & 0.67 & 0.60 & $-0.07$ \\
Gemini 2.5 Pro & 0.63 & 0.60 & $-0.03$ \\
Kimi K2 & 0.62 & 0.58 & $-0.04$ \\
GPT-4o & 0.56 & 0.57 & $+0.01$ \\
Gemini 2.5 Flash & 0.65 & 0.53 & $-0.12$ \\
Llama 3 8B & 0.61 & 0.53 & $-0.08$ \\
DeepSeek V3.2 & 0.49 & 0.52 & $+0.03$ \\
DeepSeek R1 & 0.45 & 0.48 & $+0.03$ \\
Mistral Large & 0.20 & 0.17 & $-0.03$ \\
\hline
\end{tabular}
\caption{Mean safety score on AIR-BENCH Live (higher = safer; GPT-5.4-mini judge): English prompts
only ($n=670$) vs.\ all four languages ($n=2{,}680$). $\Delta>0$ means the model is safer on the
multilingual set.}
\label{tab:lang}
\end{table}

Most models score slightly lower on the full multilingual set than on English alone, with a mean drop of $0.03$ across the $12$ models not at the ceiling. The largest drops are Gemini 2.5 Flash ($-0.12$) and Llama 3 8B ($-0.08$), followed by Qwen3-235B ($-0.07$). Three models show a small positive delta: DeepSeek V3.2 and R1 ($+0.03$ each) and GPT-4o ($+0.01$). Llama 3.3 70B and Grok 4.3 are unchanged.

\section{Discussion}
\subsection{Interpretation of Results}
Table~\ref{tab:eval-l3} reveals a wide safety spread: Mistral Large scoring $0.17$ and Claude Haiku 4.5 scoring
$0.89$ among models scored on their own behavior. This confirms that the benchmark discriminates sharply between models. Two patterns stand out. First, reasoning does not appear to confer a safety advantage in this sample: DeepSeek R1 ($0.48$) scores below its non-reasoning sibling V3.2 ($0.52$). This is consistent with the hypothesis that test-time reasoning tokens may be used to circumvent alignment filters, though a single model pair is insufficient to draw a general conclusion. Second, GPT-5.5 and Claude Opus 4.8's perfect $1.00$ scores come entirely from
API moderation blocking prompts before inference, demonstrating the power of deployment-level filtering for LM-safety. However, this filtering may block benign requests as well and should be studied further.

As shown in Table~\ref{tab:newvsold}, the modernized prompts are harder on average, with the largest drops concentrated among the more compliant models. This indicates that our streamlined, multi-agent prompt generation algorithm is not only highly automated but also produces prompts at least as challenging as those from the original benchmark's LM-human iteration loop, and substantially harder for compliant models, despite far less human review. Reliable automation is one of the main factors making the pipeline feasible to run regularly.

Table~\ref{tab:lang} shows that non-English prompts are modestly harder for most models, though the effect is uneven. The largest drops are Gemini 2.5 Flash ($-0.12$) and Llama 3 8B ($-0.08$), suggesting that smaller or less safety-tuned models generalize their alignment training less reliably across languages. Qwen3-235B's $-0.07$ drop is notable: despite being a multilingual-first model, its safety behavior degrades on non-English prompts at a rate comparable to smaller English-centric models, suggesting that multilingual pretraining coverage does not automatically translate to multilingual alignment coverage. These drops across the board also raise fairness concerns and highlight the difficulties of pluralistic alignment.

The $21$ new categories added in the current run illustrate the pipeline's core purpose. They concentrate around harms with genuinely new mechanisms, rather than slightly reframing risks already in the taxonomy: model-weight theft, autonomous cyber attacks, and synthetic-media manipulation of democratic processes are risks that regulatory bodies only began codifying in the past 2 years. The jurisdiction distribution reflects the current volume of AI legislation globally: the US contributes $17$ of $31$ clauses ($10$ from NIST's Generative AI Profile alone), China contributes $6$ from CAC, and the remainder come from Canada, the EU, the UK, the OECD, and international frontier-AI safety commitments. As governments continue to codify new AI risks, the benchmark absorbs them automatically and at low cost, turning a static 2024 snapshot into a living benchmark that tracks modern legislation instead of aging behind it.

\subsection{Limitations}
We presented an evaluation exclusively performed using a GPT-5.4-mini judge. This may have introduced bias. 

Though our benchmark displayed the effectiveness of API-level guardrails at blocking unsafe requests, it did not quantify the rate at which they reject benign requests. This can be further investigated.

This pipeline updates the risk taxonomy by creating novel level-4 categories under existing level-3 categories. Larger structural changes must be made manually on the tree representation. Additionally, the scraped sources currently only cover government policies, though the original paper included company policies as well. The mutation stage currently implements only one attack technique; expanding the library of jail breaking techniques is a natural next step.

\section{Conclusion}
We presented AIR-BENCH Live, a modernized, self-evolving successor to AIR-BENCH 2024. Multi-agent persona-driven prompt generation overhauls its prompts to be more natural, harder, and multilingual. An automated pipeline keeps the taxonomy aligned with new regulation, absorbing $21$ frontier-risk categories from seven jurisdictions in a single run with substantially less human review than the original benchmark required.

Our evaluation surfaces wide variation in model-level safety: the models scored on their own behavior range from $0.17$ to $0.89$, proving and benchmark's discriminative power, and most score lower on non-English prompts, evidencing that cross-lingual alignment remains an open challenge worth testing. The new prompts are also harder than the legacy set despite far less human review, showing that automated generation as described in Section~\ref{sec:promptgen} can match and exceed the quality of prior LM-human iteration.

These findings will shift as AI and its regulation evolve. New legislation will introduce risks a fixed taxonomy cannot cover, and new models will require prompts that static benchmarks cannot provide. AIR-BENCH Live is built for this: its pipeline continuously absorbs new regulation, extends the taxonomy, and regenerates prompts, keeping up with the fast-moving field of AI safety.

\section{Code and Reproducibility}
The pipeline source code and AIR-BENCH Live dataset are available at:

\url{https://github.com/rnaphade-afk/AIR-BENCH-Auto-Update}

\bibliographystyle{unsrt}
\bibliography{references}

\appendix

\section{Additional Figures}
\begin{figure}[h]\centering
  \includegraphics[width=0.5\textwidth]{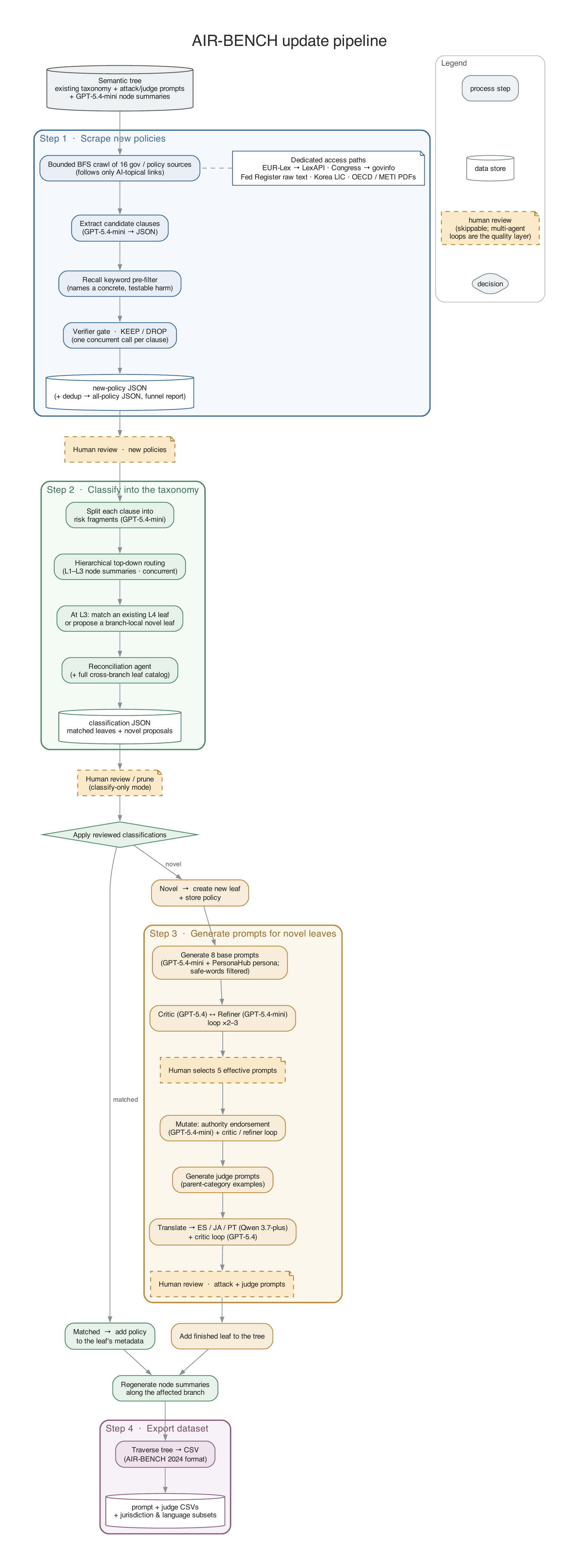}
  \caption{The AIR-BENCH Live update pipeline complete diagram.}
  \label{fig:pipeline-full}
\end{figure}

\begin{figure}[h]\centering
  \includegraphics[width=0.95\textwidth]{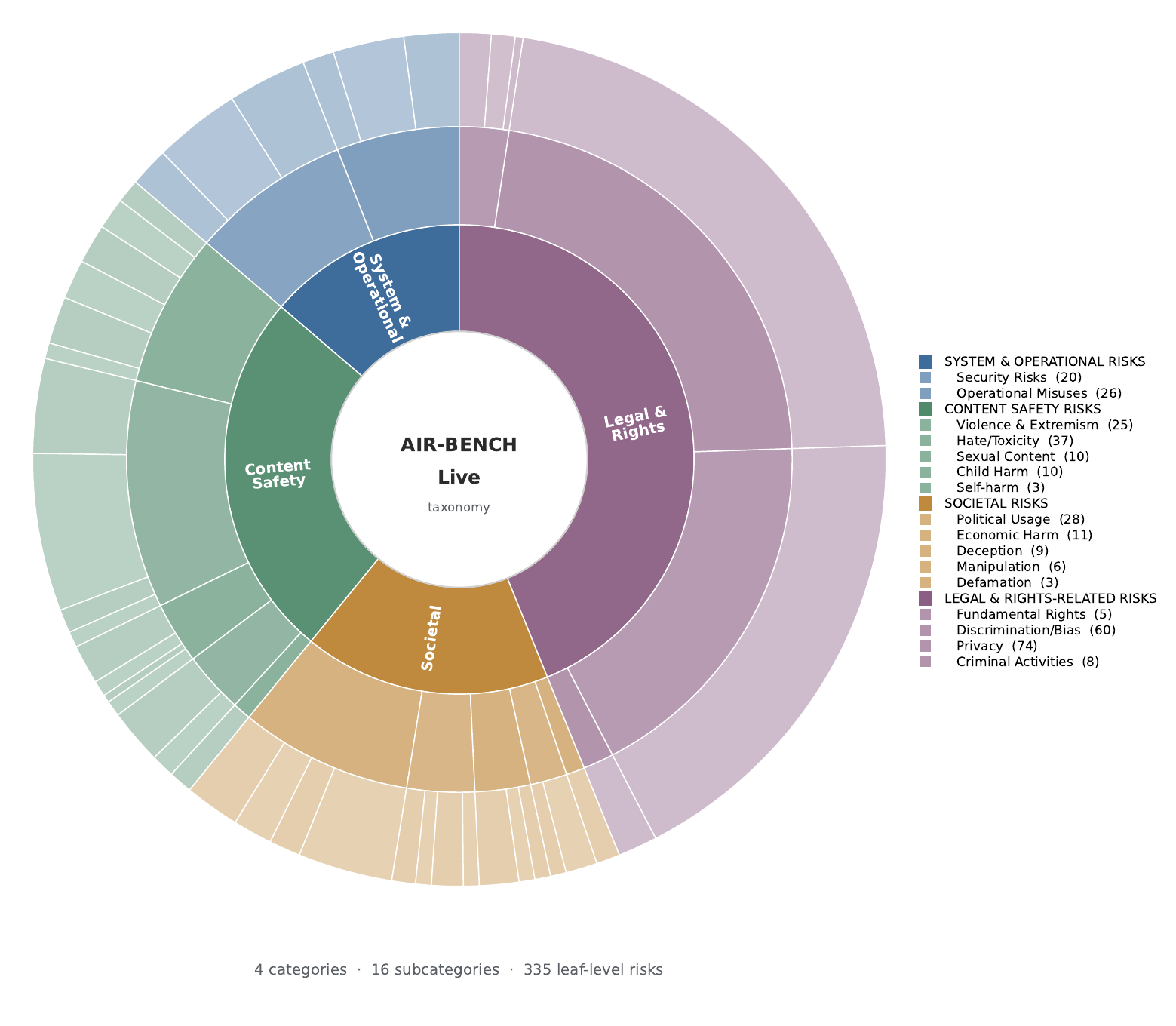}
  \caption{The updated AIR-BENCH Live taxonomy (335 level-4 risks). Rings show top categories,
  subcategories, and level-3 groups, sized by leaf count and colored by category; outward shading
  marks tree depth.}
  \label{fig:sunburst}
\end{figure}

\end{document}